%% file: root.tex
\documentclass[letterpaper,10pt,conference]{ieeeconf}
\IEEEoverridecommandlockouts
\usepackage{times}
\usepackage{graphicx}
\usepackage{amsmath,amssymb,bm}
\usepackage{booktabs,multirow,array}
\usepackage{makecell}
\usepackage{cite}
\usepackage{url}
\usepackage{xcolor}
\usepackage{placeins}
\usepackage[nocheckfootnote]{flushend}
\usepackage[hidelinks]{hyperref}
\hypersetup{
  pdftitle={Conditional Trajectory Peaks: Single-Pass Multimodal Policies over Action Chunks},
  pdfauthor={Di Wu; Rongtian Shen; Ping Liu; Xuhua Chen; He Zheng; Lingfeng Zhang; Tao Zhang},
  pdfsubject={arXiv preprint}
}
\newcommand{\ours}{\textbf{CTP}}

\title{\LARGE \bf
Conditional Trajectory Peaks: Single-Pass Multimodal Policies over Action Chunks
}
\author{Di Wu\textsuperscript{1,2}\authorrefmark{1},
Rongtian Shen\textsuperscript{1}\authorrefmark{1},
Ping Liu\textsuperscript{1}, Xuhua Chen\textsuperscript{1},
He Zheng\textsuperscript{1}, Lingfeng Zhang\textsuperscript{1},
and Tao Zhang\textsuperscript{1}\authorrefmark{2}%
\thanks{\textsuperscript{1}Magiclab Robotics Technology Co., Ltd., China.}%
\thanks{\textsuperscript{2}Southeast University, Nanjing, China.}%
\thanks{\authorrefmark{1}These authors contributed equally.}%
\thanks{\authorrefmark{2}Corresponding author.}%
}

\begin{document}
\bstctlcite{ICRA2027:reference-style}
\maketitle
\thispagestyle{empty}
\pagestyle{empty}

\input{sections/abstract}
\input{sections/introduction}
\input{sections/related_work}
\input{sections/method}
\input{sections/experiments}
\input{sections/results}
\input{sections/conclusion}

\bibliographystyle{IEEEtran}
\bibliography{references,bibliography_style}
\end{document}

%% file: sections/abstract.tex
\begin{abstract}
Multimodal imitation learning requires diverse executable futures under the same observation and consistent behavior across replanning cycles. We present Conditional Trajectory Peaks (CTP), a single-pass policy framework that jointly predicts complete action-chunk candidates, probability masses, and trajectory scales. Distribution-Aware Peak Specialization (DAPS) specializes trajectory peaks using trajectory-level posterior responsibilities and mass- and scale-modulated overlap constraints. Evidence-Gated Trajectory Belief Transport (ETBT) maintains cross-chunk consistency through geometric correspondence between exchangeable candidate sets, while allowing current policy evidence to override historical constraints. CTP achieves a coverage score of 91.40\% on Push-T; success rates of 100.0\%, 79.72\%, and 84.44\% on D3IL Avoiding, Aligning, and Sorting-2, respectively. On LIBERO, CTP achieves an average success rate of 97.25\%. In real-world dual-arm experiments, CTP preserves both placement modes in a two-plate task, succeeding in all 50 trials. On bottle uprighting and pen placement into a holder, it maintains success rates comparable to \(\pi_{0.5}\) while reducing policy inference latency from 218.24~ms to 75.80~ms. These results demonstrate that single-pass trajectory modeling can combine multimodal behavior, closed-loop consistency, and efficient inference.
\mbox{\textbf{Project Page:}~\href{https://embodied.magiclab.top/works/ctp/index.html}{\textcolor{blue}{\textit{CTP}}}}
\end{abstract}

{\small\noindent\textbf{Keywords---}Multimodal imitation learning; structured trajectory modeling; action chunking; single-pass inference; mode preservation\par}

%% file: sections/introduction.tex
\section{Introduction}

Robotic manipulation demonstrations are often multimodal: similar observations can admit different obstacle-avoidance paths, contact locations, pushing sequences, or subtask orders. Policies must represent these distinct executable futures rather than a single deterministic action. Squared-error behavioral cloning instead favors the conditional mean, which can enter low-probability or infeasible regions when trajectory modes are separated\cite{bishop1994mdn}. Diffusion Policy, BESO, and flow-matching policies model complex multimodal action distributions\cite{chi2023diffusion,reuss2023beso,lipman2023flow,physicalintelligence2025pi05}. However, diffusion denoising and numerical flow integration generally require multiple network evaluations per replanning cycle. This iterative generation increases latency and computation in high-frequency closed-loop control. This raises a central question: \emph{Can a policy generate trajectory-level multimodal actions with one network forward pass per replanning step?}

\begin{figure}[!t]
\centering
\includegraphics[width=\columnwidth]{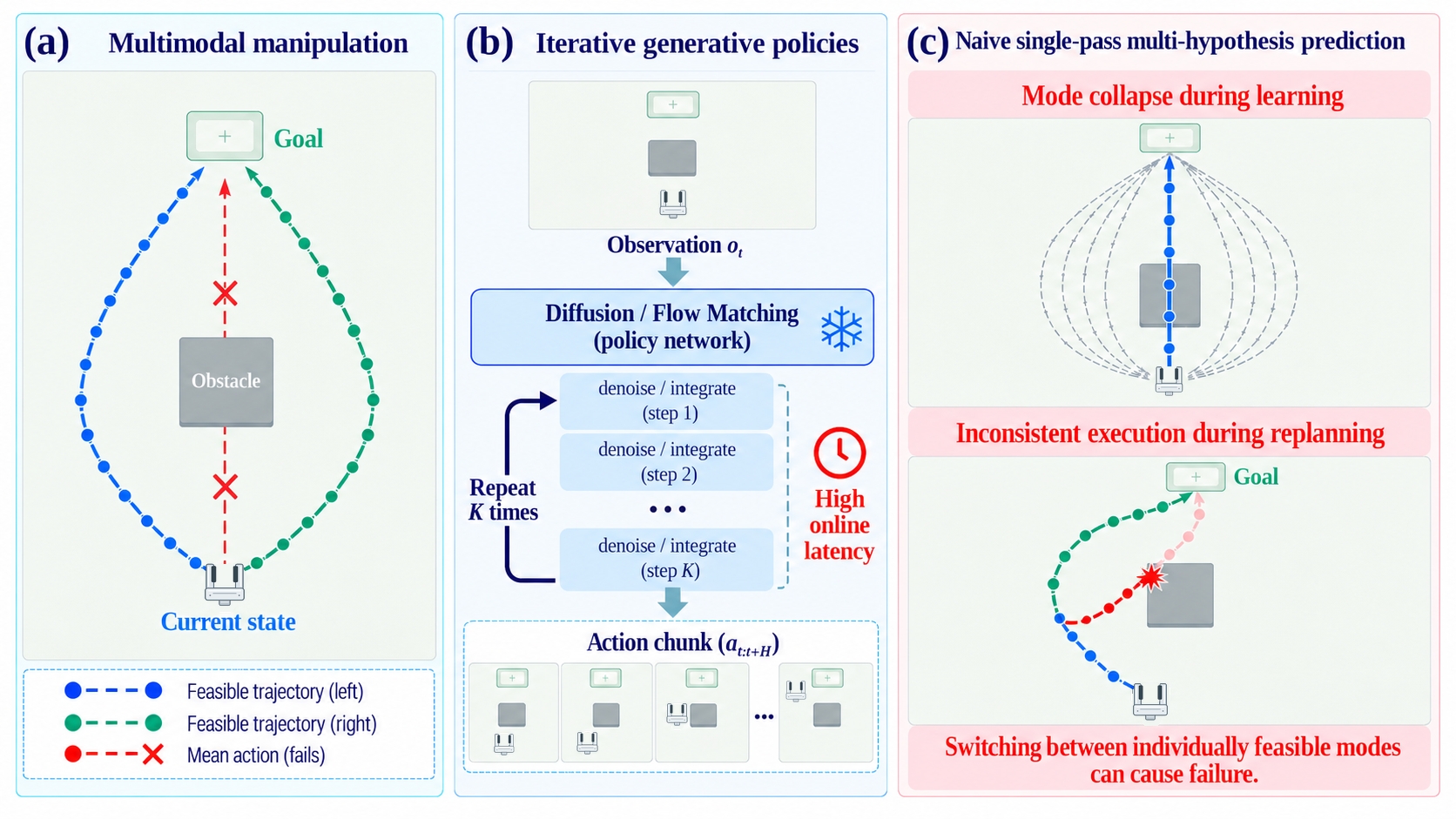}
\caption{Core challenges in multimodal closed-loop manipulation. (a) The same observation admits multiple feasible futures, while the conditional mean may be infeasible. (b) Diffusion and flow matching policies generate action chunks iteratively, increasing online decision latency. (c) Naive single-pass multi-candidate prediction can still suffer from mode collapse during training and inconsistent execution caused by mode switching during receding-horizon replanning.}
\label{fig:intro-motivation}
\end{figure}

A direct alternative is to predict multiple trajectories in parallel. Existing multi-hypothesis policies demonstrate single-pass batched candidate generation\cite{rana2025imle,bhaskar2026prism}, but parallel outputs do not guarantee effective multimodality. Permutation symmetry can drive candidates toward the same dominant high-density region early in training. This produces \emph{mode collapse}: nominally distinct candidates represent identical or similar behaviors, leaving the model effectively unimodal. Additional output capacity therefore permits multiple modes without ensuring allocation to distinct data-supported regions. Fixed action anchors, offline clustering, or predefined mode quotas impose candidate roles through data processing or model design. Single-pass policies instead need data-driven specialization that assigns high-probability trajectories to distinct valid modes and reduces probability mass on unsupported redundant candidates.

Candidate diversity alone is insufficient for multimodal closed-loop control. Action-chunking policies execute only a chunk's first few steps before replanning, so candidates must preserve behavioral continuity across decisions. Local probability changes can switch the policy between incompatible routes, stitching individually feasible chunks into discontinuous or unsuccessful behavior, a failure we call \emph{cross-chunk mode stitching}. Unlike training-time mode collapse, stitching concerns behavioral identity across replanning cycles and can occur despite diverse candidates. Because finite mixture components are permutation-symmetric, their indices cannot identify corresponding behaviors across successive exchangeable candidate sets. We therefore formulate cross-chunk consistency as \emph{behavioral identity correspondence and sequential transfer}: physical trajectory relationships identify continuations, while current policy evidence can revise or overturn earlier choices. Fig.~\ref{fig:intro-motivation} summarizes these challenges.

We therefore propose \textbf{Conditional Trajectory Peaks (CTP)}, a single-pass policy framework for multimodal closed-loop control. CTP models trajectory-level conditional peaks over complete action chunks. One shared network forward pass provides multimodal prediction with $\mathrm{NFE}=1$, jointly addressing peak specialization and behavioral consistency across replanning cycles.

Our main contributions are as follows:

\begin{itemize}

\item We propose CTP, a single-pass framework for multimodal closed-loop control that jointly predicts multiple trajectory candidates, probabilities, and scales with $NFE=1$.

\item We introduce Distribution-Aware Peak Specialization (DAPS), which mitigates mode collapse through posterior responsibility assignment and probability- and scale-aware inter-peak regularization, without clustering or predefined mode assignments.

\item We introduce Evidence-Gated Trajectory Belief Transport (ETBT), which improves cross-chunk behavioral consistency by propagating beliefs across replanning cycles while adaptively revising them using current policy evidence.

\end{itemize}

%% file: sections/related_work.tex
\section{Related Work}

\subsection{Multimodal Policies for Robotic Imitation Learning}

Squared-error behavioral cloning tends toward the conditional mean and struggles with multivalued action mappings. Mixture density networks (MDNs) predict mixture parameters, providing a classical probabilistic framework for continuous multimodal regression\cite{bishop1994mdn}. With vectorized action chunks, a trajectory-level MDN uses one discrete component to explain the entire trajectory and optimizes a normalized mixture likelihood in one forward pass.

IBC represents implicit conditional distributions through an energy function and approximates inference by sampling actions\cite{florence2022ibc}. BeT clusters continuous actions into discrete categories and predicts cluster indices and continuous residuals\cite{shafiullah2022bet}. ACT shortens the effective prediction horizon through action chunking\cite{zhao2023act}, while D3IL systematically evaluates how different policies learn diverse human demonstrations\cite{jia2024d3il}. Energy Policy trains a noise-conditioned energy network to generate complete action sequences in parallel without iterative denoising\cite{jia2025energy}. Diffusion Policy and BESO demonstrate the effectiveness of iterative generation for multimodal robot control\cite{chi2023diffusion,reuss2023beso}. These approaches improve distributional expressiveness, temporal modeling, or generation quality, but trade-offs remain among multimodal representation, policy inference overhead, and closed-loop candidate selection.

\subsection{Single-Pass Multi-Hypothesis Generation}

IMLE Policy regresses the demonstration's closest candidate among action sequences generated from random latent variables, and uses the previous chunk's unexecuted suffix for consistency-based selection\cite{rana2025imle}. PRISM combines full-batch rejection-sampling IMLE, Performer, and learned action queries to generate multiple sequences in one batched forward pass\cite{bhaskar2026prism}. Liquid Networks with Mixture Density Heads predicts a Gaussian mixture per time step\cite{correll2026liquid}. Designs primarily trade off candidate sampling, learned action queries, and per-step mixtures. Direct parameterization of normalized complete action-chunk distributions in one shared forward pass, with one latent component spanning the entire horizon, remains insufficiently studied.

\subsection{Consistency and Feedback across Action Chunks}

BID samples multiple action chunks at inference, using backward coherence to preserve consistency with past decisions and forward contrast to guide candidate selection\cite{liu2025bid}. IntentVLA extracts short-horizon intent from recent visual history and conditions action-chunk generation on it, mitigating intent conflicts under observation aliasing\cite{lian2026intentvla}. IMLE Policy also uses the unexecuted suffix of the previous action chunk for consistency-based selection\cite{rana2025imle}. These studies highlight the importance of maintaining prior action intent. However, identifying the same physical route within candidate sets with permutable component indices, and actively releasing constraints when current evidence contradicts history, remain key challenges in receding-horizon execution.

\subsection{Multimodal Trajectory Modeling in Motion Prediction}

Motion prediction also represents multiple futures from limited history. AutoBots uses latent-variable set Transformers for single-pass scene-consistent trajectory prediction\cite{girgis2022autobots}; MotionDiffuser models joint multi-agent trajectory distributions through diffusion\cite{jiang2023motiondiffuser}. Multi-hypothesis regression commonly uses winner-takes-all objectives; aWTA stabilizes optimization through annealing\cite{xu2025awta}, while ModeSeq explicitly models sequential relationships among modes\cite{zhou2025modeseq}. Robotic closed-loop control shares trajectory-level multimodality with motion prediction but additionally handles receding-horizon replanning, contact feedback, and temporal permutations of finite mixture components. Single-step candidate diversity therefore cannot fully characterize closed-loop performance.

%% file: sections/method.tex
\begin{figure*}[!t]
\centering
\includegraphics[width=\textwidth]{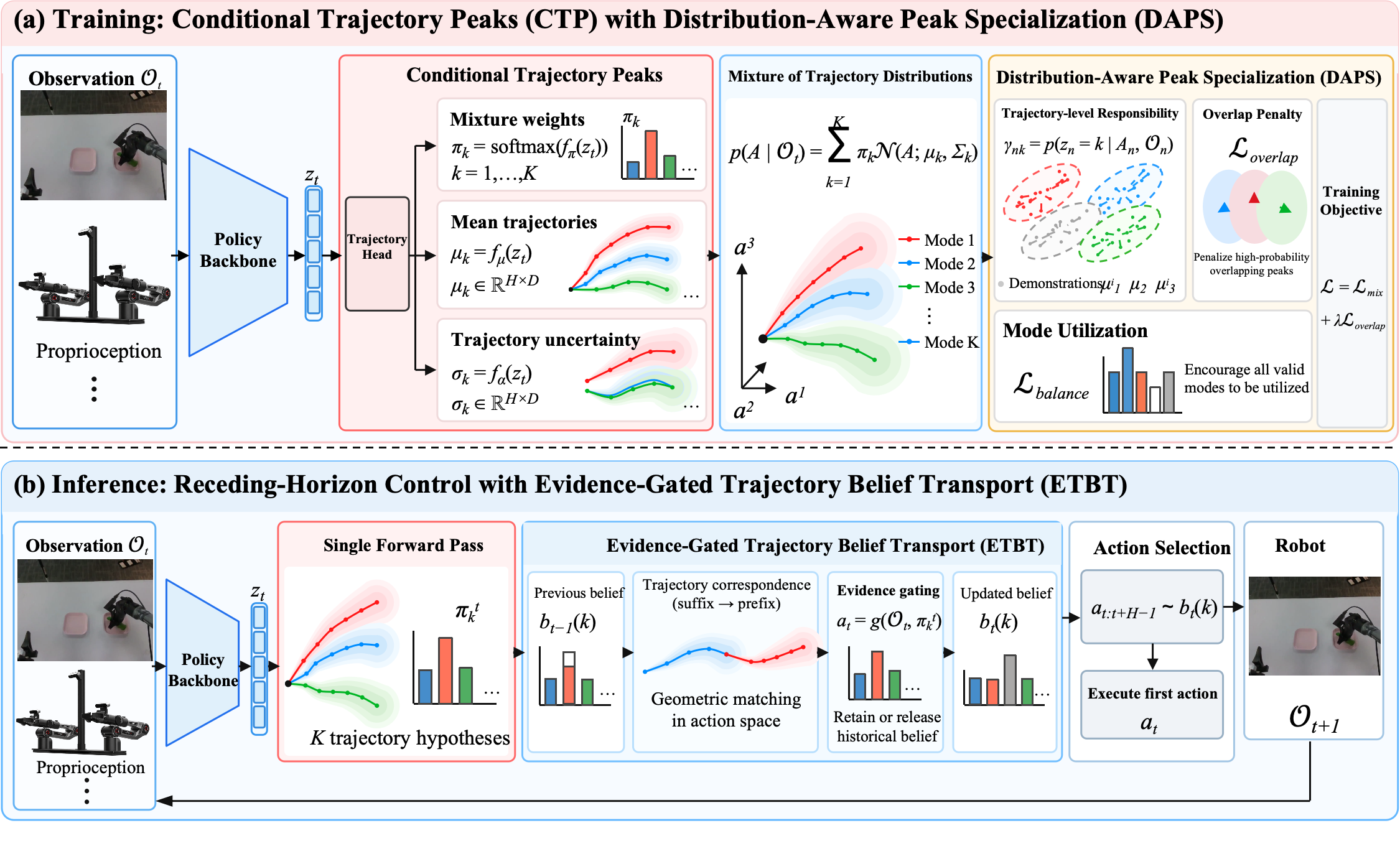}
\caption{Overview of CTP. (a) Training: the observation history passes through a shared policy backbone and trajectory head once to generate $K$ trajectory peaks with probability masses and uncertainty. DAPS promotes mode specialization through trajectory-level responsibility assignment and inter-peak overlap constraints. (b) Inference: ETBT propagates historical behavioral beliefs through geometric correspondences between candidate trajectories in successive replanning cycles. Current policy evidence determines whether historical constraints are retained or released.}
\label{fig:ctp-model-architecture}
\label{fig:method}
\end{figure*}

\section{Method}

\subsection{Problem Formulation and Framework Overview}

Given offline demonstrations \(\mathcal{D} = \{(\mathcal{O}_t,A_t)\},\) let $\mathcal{O}_t=[o_{t-L+1},\ldots,o_t]$ denote the length-$L$ observation history and $A_t=[a_t,\ldots,a_{t+H-1}]\in\mathbb{R}^{H\times D}$ the target action chunk with horizon $H$ and action dimension $D$. We seek a conditional policy \(p_\theta(A_t\mid\mathcal{O}_t)\) representing distinct executable futures under the same observation.

Tasks may use absolute positions, relative displacements, or other action parameterizations. For trajectory modeling, we use the normalized representation \(\widetilde{A}_t = \Psi_a(A_t;q_t)\), where $q_t$ is the current action reference state and $\Psi_a$ performs task-specific normalization and reference transformation. This unifies trajectory representations without predefining behavioral modes.

Our single-pass framework for multimodal closed-loop control, CTP, comprises three modules (Fig.~\ref{fig:ctp-model-architecture}): trajectory-level conditional peak modeling, DAPS, and ETBT. One shared backbone forward pass predicts complete trajectory candidates, probability masses, and scales. DAPS specializes peaks during training; ETBT transfers behavioral beliefs across replanning cycles. Each module is detailed below.

\subsection{Trajectory-Level Conditional Peak Modeling}

The shared policy backbone first encodes the observation history into a context representation
\(z_t = f_\theta(N_o(\mathcal{O}_t)) \in\mathbb{R}^{M}.\)
The trajectory parameter head generates all $K$ trajectory peaks in one shared forward pass. For peak $k$, it predicts trajectory parameters $C_{t,k}\in\mathbb{R}^{R\times D}$, a probability logit $\ell_{t,k}$, and a scale parameter $s_{t,k}$. Given a fixed trajectory basis $B\in\mathbb{R}^{H\times R}$, we define
\begin{equation}
\mu_{t,k}=BC_{t,k},
\qquad
\pi_t=\operatorname{softmax}(\ell_t),
\end{equation}
and
\begin{equation}
\sigma_{t,k}
=
\sigma_{\min}
+
(\sigma_{\max}-\sigma_{\min})
\operatorname{sigmoid}(s_{t,k}).
\end{equation}

Here, $\mu_{t,k}$ is the complete trajectory center of mode $k$, and $\pi_{t,k}=p(Z=k\mid\mathcal{O}_t)$ its probability mass under the current observation. The trajectory scale $\sigma_{t,k}$ describes conditional residual dispersion around that center in normalized action-chunk space. In particular, $\pi_{t,k}$ is not execution success probability, nor is $\sigma_{t,k}$ physical path width.

Using these parameters, CTP defines a finite conditional mixture over complete action chunks:
\begin{equation}
p_\theta(\widetilde{A}\mid\mathcal{O})
=
\sum_{k=1}^{K}
\pi_k(\mathcal{O})
\mathcal{N}
\left(
\operatorname{vec}(\widetilde{A});
\operatorname{vec}(\mu_k),
\sigma_k^2 I_{HD}
\right).
\end{equation}
The discrete variable $Z\in\{1,\ldots,K\}$ selects one complete trajectory component to jointly explain the entire horizon. Mode identity therefore belongs to the complete action chunk, rather than varying independently at each step.

We call each component center $\mu_k$ a \emph{trajectory peak}. The learned scale $\sigma_k$ captures within-mode residual dispersion: larger residuals favor broader components, whereas the likelihood penalizes unrestricted inflation.

The basis $B$ supports low-rank action-chunk parameterization ($R<H$, using a low-frequency basis) or direct time-domain prediction ($R=H$, $B=I_H$). Benchmark-specific choices appear in the experimental setup. Both retain a shared forward pass and $NFE=1$ per replanning cycle.

\subsection{DAPS}

Finite mixtures provide capacity for multiple future behaviors, but more candidates alone do not ensure coverage of distinct data modes. Early in training, exchangeable candidates may converge to the demonstration distribution's dominant high-density region, producing nominally multiple but redundant peaks. DAPS addresses this mode collapse.

Let $d=HD$. For demonstration action chunk $n$ and trajectory peak $k$, the Gaussian energy is
\begin{equation}
E_{nk}
=
\frac{
\|\widetilde{A}_n-\mu_{nk}\|_F^2
}{
2\sigma_{nk}^2
}
+
d\log\sigma_{nk}
+
\frac{d}{2}\log(2\pi).
\end{equation}
Marginalizing the discrete trajectory variable gives the mixture negative log-likelihood
\begin{equation}
\mathcal{L}_{\mathrm{mix}}
=
-\frac{1}{N}
\sum_{n=1}^{N}
\log
\sum_{k=1}^{K}
\exp
\left(
\log\pi_{nk}-E_{nk}
\right).
\end{equation}
The corresponding trajectory-level posterior responsibility is
\(r_{nk} = \operatorname{softmax}_{k} \left( \log\pi_{nk}-E_{nk} \right).\)

Unlike hard winner-takes-all assignment based solely on trajectory distance, $r_{nk}$ jointly considers fitting error, probability mass, and trajectory scale. It allocates learning signals consistently with the conditional mixture distribution without predefined candidate-to-mode mappings.

Soft responsibility assignment alone, however, does not ensure effective specialization. To further discourage high-probability peaks from repeatedly explaining the same data region, we define the normalized trajectory distance
\begin{equation}
d_{n,ij}
=
\frac{
\|\mu_{ni}-\mu_{nj}\|_F^2
}{
HD
},
\end{equation}
and construct an inter-peak overlap constraint
\begin{equation}
\mathcal{L}_{\mathrm{ov}}
=
\frac{1}{N}
\sum_{n=1}^{N}
\sum_{i<j}
\pi_{ni}\pi_{nj}
\exp
\left[
-\frac{
d_{n,ij}
}{
2(\sigma_{ni}^2+\sigma_{nj}^2)
}
\right].
\end{equation}

This constraint induces distribution-aware selective specialization. The probability factor $\pi_{ni}\pi_{nj}$ concentrates separation pressure on pairs with high probability mass, while the scale term assesses overlap relative to their distributional widths. Unlike uniform repulsion in generic diversity regularization, DAPS suppresses data-supported, high-probability peaks that are mutually redundant. The overlap constraint encourages separation of independently supported candidates into distinct modes. Unsupported redundant candidates can reduce their probability mass and contribution, without forced assignment to artificially created additional modes.

The core DAPS training objective is
\begin{equation}
\mathcal{L}_{\mathrm{DAPS}}
=
\mathcal{L}_{\mathrm{mix}}
+
\lambda_{\mathrm{ov}}
\mathcal{L}_{\mathrm{ov}}.
\end{equation}
Trajectory-level posterior responsibilities assign candidate roles according to their ability to explain the data, while the distribution-aware overlap constraint discourages high-probability peaks from persistently explaining the same region. Together, they promote data-driven peak specialization. Backbone-specific auxiliary terms for candidate calibration or training stabilization appear in the experimental setup, outside the core DAPS definition.

\subsection{Single-Pass Decoding and Receding-Horizon Execution}

Given the current observation, CTP obtains all trajectory peaks in one forward pass. Without cross-cycle history, it uses maximum-probability decoding:
\(k_t^{\mathrm{raw}} = \arg\max_k \pi_{t,k}.\)
The trajectory centers are mapped back to the task action space as
\(\widehat{A}_{t,k} = \Psi_a^{-1}(\mu_{t,k};q_t).\)
The policy executes the first $E\le H$ actions of the selected chunk, then replans from new observations. Candidate generation requires no additional sampling, iterative denoising, or vector-field integration, so the procedure maintains $NFE=1$.

\subsection{ETBT}

ETBT maintains behavioral identity during receding-horizon execution. Rather than merely smoothing actions, it explicitly matches behavior across trajectory sets with permutable indices and fuses historical behavior as a soft prior with current policy evidence.

First, a task-specific mapping $\Phi$ transforms candidate action chunk $k$ into a trajectory representation that supports geometric comparison across cycles:
\(P_{t,k} = \Phi(\widehat{A}_{t,k};q_t).\)
If $E$ steps have been executed since the previous replanning cycle, let $M=H-E$. The correspondence cost between the unexecuted suffix of previous candidate $i$ and the prefix of current candidate $j$ is
\begin{equation}
C_t(i,j)
=
\frac{1}{M D_P}
\sum_{h=1}^{M}
\left\|
P_{t-1,i,E+h}
-
P_{t,j,h}
\right\|_2^2 ,
\end{equation}
where $D_P$ is the dimension of the representation used for trajectory matching.

We normalize the correspondence cost by a robust trajectory scale $s_t$ and construct a similarity kernel
\(G_t(i,j) = \exp \left( -\frac{ C_t(i,j) }{ s_t\tau } \right),\)
then obtain conditional correspondence probabilities using a normalization operator $\Gamma(\cdot)$ that preserves equivariance to candidate permutations:
\begin{equation}
T_t(j\mid i)
=
\Gamma(G_t)_{ij},
\qquad
\sum_j T_t(j\mid i)=1.
\end{equation}
Depending on the task, this can be implemented by row-wise conditional normalization or Sinkhorn normalization with marginal constraints.

Given the behavioral belief $b_{t-1}(i)$ over the previous candidate set, trajectory correspondences propagate a belief about behavioral continuation into the current cycle:
\(c_t(j) = \sum_i b_{t-1}(i) T_t(j\mid i).\)
If only the previously executed candidate is retained, $b_{t-1}$ reduces to a one-hot distribution on that candidate.

We smooth $c_t$ into a historical behavioral prior
\(\bar b_t = \rho c_t + (1-\rho)u,\)
where $u$ is a weak fallback prior and $\rho$ controls the strength of historical behavior retention. Current policy probabilities provide evidence under the latest observation:
\(q_t(j) = \pi_t(j)^\beta .\)
Fusing the historical prior and current policy evidence yields
\begin{equation}
b_t(j)
=
\frac{
\bar b_t(j)q_t(j)
}{
\sum_m
\bar b_t(m)q_t(m)
}.
\end{equation}

The key to ETBT is that history acts only as a soft prior, rather than an inviolable hard constraint. Let
\(j_c = \arg\max_j c_t(j), \qquad j_q = \arg\max_j q_t(j).\)
If the candidate most strongly supported by the current policy has substantially stronger evidence than the candidate continuing the historical behavior, namely,
\(\frac{ q_t(j_q) }{ q_t(j_c)+\epsilon } \ge\gamma,\)
or if the cross-cycle trajectory correspondence is unreliable,
\(\max_j c_t(j)<\eta,\)
the historical constraint is released and candidate selection reverts to current policy evidence. Otherwise, the selected candidate is
\(k_t^\star = \arg\max_j b_t(j).\)

ETBT thus maintains behavioral continuation across cycles without relying on fixed candidate indices. It releases historical constraints when current evidence is sufficiently strong or trajectory correspondences are unreliable, without additional policy backbone forward passes.

%% file: sections/experiments.tex
\section{Experiments}

\subsection{Experimental Setup}

\textbf{Push-T:}
We use the Push-T closed-loop environment from Diffusion Policy\cite{chi2023diffusion}. Policies receive two historical states and predict $16\times2$ action chunks. CTP and the single-peak baseline use $K=4/1$, respectively, with the same Transformer backbone, training data, and 40-epoch training configuration. After five training runs, the final EMA weights are evaluated on 200 fixed initial conditions.

\textbf{D3IL:}
We use three state-based control tasks from D3IL: Avoiding, Aligning, and Sorting-2\cite{jia2024d3il}. Their multiple solutions arise from obstacle-avoidance paths, inner- versus outer-side alignment, and object sorting order, respectively. These tasks cover three representative multimodal control settings: path planning, pose alignment, and object sorting. They evaluate CTP's ability to model different types of multimodal behavior and its policy selection.

\textbf{LIBERO:}
We use the Spatial, Object, Goal, and Long suites of LIBERO\cite{liu2023libero}, comprising 40 tasks. CTP freezes the $\pi_{0.5}$ vision-language backbone, uses $K=4$, predicts 50-step action chunks, replans every 10 steps, and selects candidates through the belief decoder.

%% file: sections/results.tex
\subsection{Main Results on Push-T}

\begin{table}[!t]
\caption{Closed-loop performance of multimodal policies on Push-T.}
\label{tab:pusht-comparison}
\centering
\scriptsize
\setlength{\tabcolsep}{3pt}
\begin{tabular}{@{}lcc@{}}
\toprule
Method & Reported results & Unified evaluation \\
\midrule
IBC/DFO\cite{chi2023diffusion,florence2022ibc}
& $0.90/0.84$
& $0.8340$ \\

BeT\cite{chi2023diffusion,shafiullah2022bet}
& $0.79/0.70$
& $0.7817$ \\

Energy Policy\cite{jia2025energy}
& $0.85$
& $0.7300$ \\

Diffusion Policy-C\cite{chi2023diffusion}
& $0.95/0.91$
& $0.8790$ \\

IMLE Policy\cite{rana2025imle}
& $0.59/0.54$
& $0.8874$ \\

Liquid-MDN\cite{correll2026liquid}
& $0.91$
& $0.7394$ \\

Trajectory-level MDN\cite{bishop1994mdn}
& --
& $0.7963$ \\

\ours
& --
& $\mathbf{0.9140}$ \\
\bottomrule
\end{tabular}
\end{table}

\subsubsection{Comparison with Existing Multimodal Policies}

Table~\ref{tab:pusht-comparison} compares CTP with six multimodal policies: IBC/DFO, BeT, Energy Policy, Diffusion Policy-C, IMLE Policy, and Liquid-MDN, plus a standard trajectory-level MDN as a direct baseline. It includes published results and a unified evaluation. Because training and evaluation settings vary across studies, quantitative comparisons primarily use the unified results.

Under the unified protocol, CTP scores $0.9140$, the highest closed-loop performance among compared methods. The strongest prior methods, IMLE Policy and Diffusion Policy-C, score $0.8874$ and $0.8790$, respectively. CTP improves over IMLE Policy by $0.0266$ and trajectory-level MDN by $0.1177$, indicating that explicit trajectory-level candidates further improve multimodal closed-loop performance.

CTP predicts multiple trajectories with NFE$=1$, without iterative sampling of multimodal actions. These results demonstrate effective single-pass trajectory-level multimodal modeling and strong Push-T closed-loop performance.

\subsubsection{Inference Efficiency}

Single-pass inference substantially reduces CTP's online policy invocation cost. CTP with $K=4$ has $74.041$M parameters, more than the compared Diffusion Policy ($65.783$M); the latency difference therefore cannot be attributed to a smaller model. On the same RTX 5070, CTP's mean and P95 invocation latencies are $0.9507$ ms and $0.9745$ ms, respectively. Diffusion Policy with NFE$=100$ requires $902.8522$ ms and $904.8401$ ms, respectively. The efficiency difference primarily results from the number of network evaluations required for parallel single-pass output versus iterative generation.

\begin{table}[t]
\caption{Model size and policy invocation latency on the same hardware.}
\label{tab:latency}
\centering
\scriptsize
\setlength{\tabcolsep}{3pt}
\begin{tabular}{@{}lcccc@{}}
\toprule
Method & Params. (M) & NFE & Mean (ms) & P95 (ms) \\
\midrule
\ours & $74.041$ & 1 & $\mathbf{0.9507}$ & $\mathbf{0.9745}$ \\
Diffusion Policy & $65.783$ & 100 & $902.8522$ & $904.8401$ \\
\bottomrule
\end{tabular}
\end{table}

\subsection{Results on Key D3IL Tasks}

Table~\ref{tab:d3il} reports closed-loop success on three representative D3IL tasks. Avoiding, Aligning, and Sorting-2 admit multiple obstacle-avoidance paths, alignment strategies, and object sorting orders, respectively, testing different types of multimodal behavior modeling. Prior results primarily serve as performance references because training and evaluation protocols are not fully matched.

\begin{table}[!t]
\caption{Closed-loop success rates (\%) on D3IL multimodal control tasks.}
\label{tab:d3il}
\centering
\scriptsize
\setlength{\tabcolsep}{2.5pt}
\resizebox{\columnwidth}{!}{%
\begin{tabular}{@{}lcccc@{}}
\toprule
Task & BC-MLP & DDPM-MLP & Best prior result & CTP \\
\midrule
Avoiding  & $66.6\pm51.2$ & $63.7\pm5.5$ & BESO $95.0\pm1.5$ & $\mathbf{100.0}$ \\
Aligning  & $70.8\pm5.2$  & $76.3\pm3.9$ & VAE-ACT $\mathbf{89.1\pm2.2}$ & $79.72$ \\
Sorting-2 & $44.4\pm6.9$  & $46.0\pm3.9$ & DDPM-ACT $\mathbf{88.2\pm2.3}$ & $84.44$ \\
\bottomrule
\end{tabular}
}
\end{table}

CTP achieves success rates of $100.0\%$, $79.72\%$, and $84.44\%$ on Avoiding, Aligning, and Sorting-2, respectively. Its $100.0\%$ success rate on Avoiding exceeds the best previously reported result of $95.0\%$. On Sorting-2, CTP achieves $84.44\%$, close to the best prior result of $88.2\%$.

\begin{figure*}[!t]
\centering
\includegraphics[width=\textwidth]{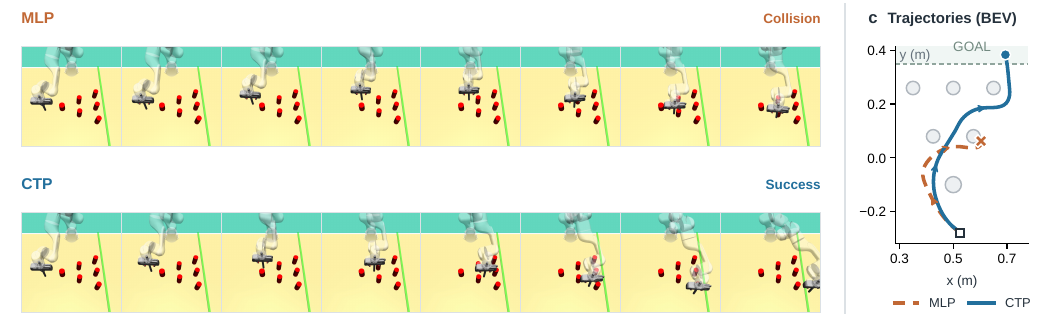}
\caption{Qualitative comparison on D3IL Avoiding. The deterministic MLP trained with MSE (top) and full CTP (bottom) start from the same initial state. Each row presents eight frames spanning a complete rollout; the bird's-eye view (BEV) shows the corresponding executed trajectories. In this example, the MLP collides with an obstacle, while CTP successfully reaches the goal.}
\label{fig:avoiding-qualitative}
\end{figure*}

In the example shown in Fig.~\ref{fig:avoiding-qualitative}, starting from the same initial state, the deterministic MLP collides with an obstacle, whereas CTP follows a collision-free route to the goal during closed-loop execution.

On Aligning, we test candidate count under strictly matched training and evaluation. Increasing $K=1$ to $K=4$ with all else fixed raises success from $69.17\%$ to $79.72\%$: a $10.56$ percentage-point gain, with a stratified $95\%$ confidence interval of $[0.28, 21.11]$ percentage points. Greater trajectory candidate capacity thus improves multimodal behavior modeling and closed-loop execution on Aligning.

Overall, CTP achieves competitive closed-loop performance across tasks with distinct multimodal structures in path selection, alignment strategy, and execution order.
\input{sections/libero_results}


\begin{figure*}[!t]
\centering
\includegraphics[width=\textwidth]{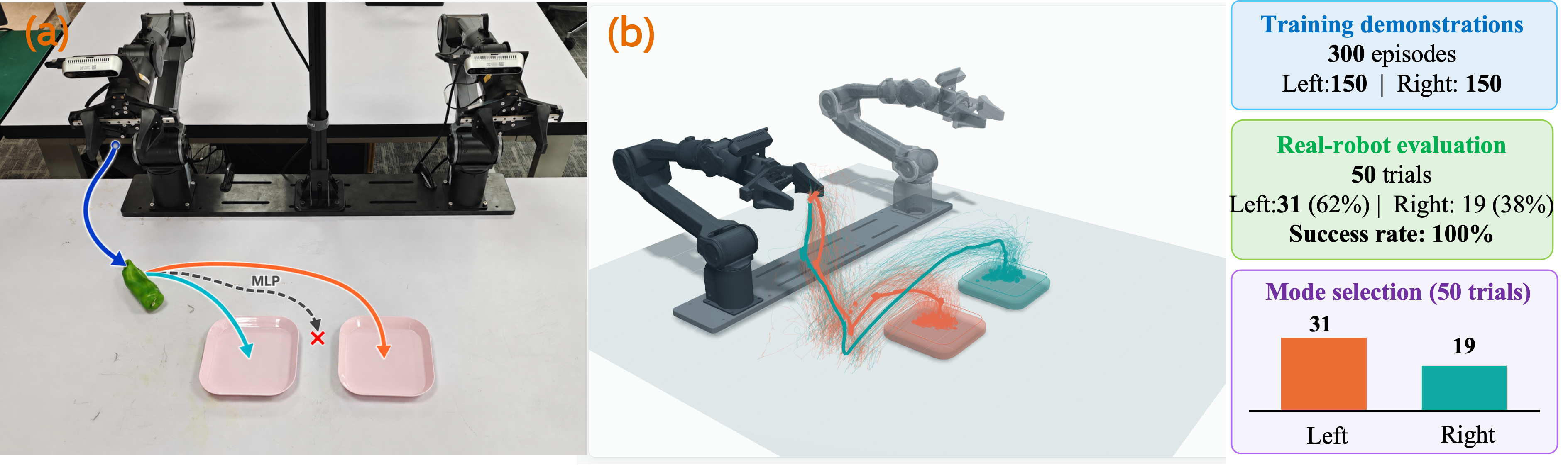}
\caption{Placement into either of two plates. (a) The real-world setup. After grasping the target object, the robot may place it in either the left or right plate; both outcomes count as success. (b) Demonstration trajectories visualized in a 3D scene model. Orange and cyan trajectories correspond to the left- and right-plate placement modes, respectively. We collected 300 demonstration episodes, with 150 per mode, yielding a balanced, spatially separated bimodal trajectory distribution.}
\label{fig:real-two}
\end{figure*}

\begin{figure*}[!t]
\centering
\includegraphics[width=\textwidth]{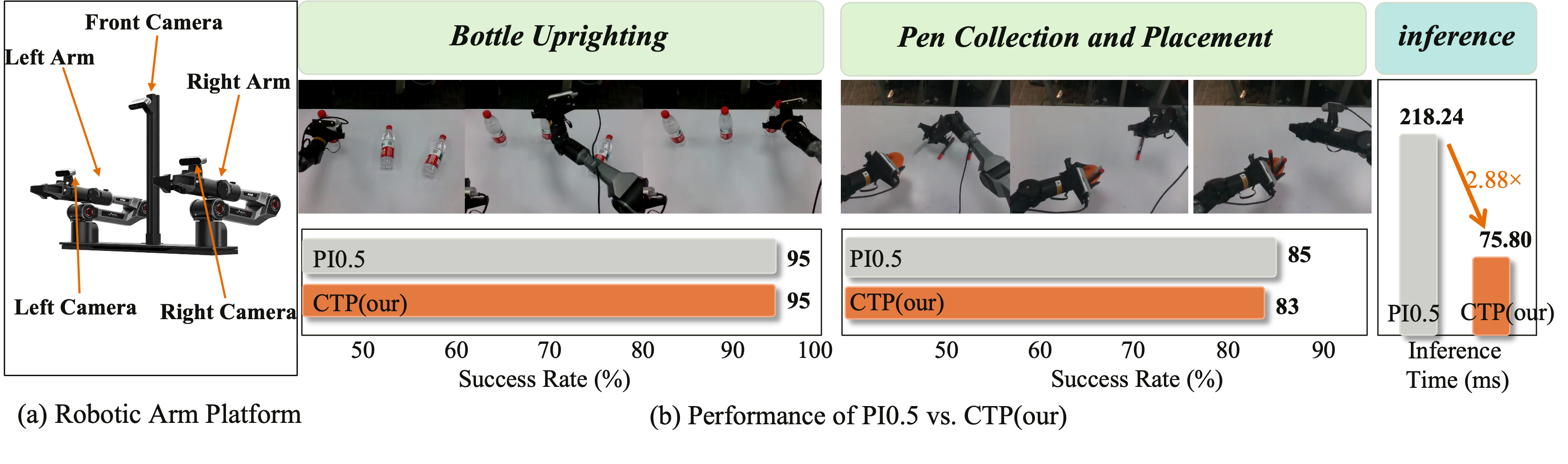}
\caption{Real-world bottle uprighting and pen placement experiments and comparative results. (a) Dual-arm robot platform and camera configuration. (b) Representative execution sequences, task success rates, and per-call policy inference latencies of $\pi_{0.5}$ and CTP on the same NVIDIA RTX 4090 GPU.}
\label{fig:real-robot-comparison}
\end{figure*}

\subsection{Ablations of Key Components}

\begin{table}[t]
\caption{Incremental component ablations of CTP on Push-T, measured by Mean Score (\%).}
\label{tab:daps-ablation}
\centering
\footnotesize
\setlength{\tabcolsep}{3pt}
\begin{tabular}{lc}
\toprule
Configuration & Mean Score (\%) \\
\midrule
Single-peak baseline ($K=1$) & $77.34$ \\
Multi-peak baseline ($K=4$) & $80.30$ \\
Multi-peak + DAPS & $85.67$ \\
CTP (full) & $\mathbf{91.40}$ \\
\bottomrule
\end{tabular}
\end{table}

To analyze each component's contribution, we start from a single-peak baseline and successively introduce trajectory-level multimodal representation, DAPS, and ETBT (Table~\ref{tab:daps-ablation}). All configurations follow the training and evaluation protocol of the main experiments.

Increasing $K$ from 1 to 4 raises Mean Score from $77.34\%$ to $80.30\%$, a gain of $2.96$ percentage points. This indicates that trajectory-level multimodal representation improves coverage of multimodal behavior. Adding DAPS further increases performance to $85.67\%$, a gain of $5.37$ percentage points, indicating that peak specialization reduces candidate redundancy and promotes differentiation among modes.

Finally, enabling ETBT for closed-loop candidate selection with the same model weights raises Mean Score to $91.40\%$, a further gain of $5.73$ percentage points. This shows that, beyond obtaining candidate trajectories with good mode coverage, maintaining consistent behavioral choices during receding-horizon replanning is also key to improving closed-loop performance.

\subsection{Real-World Robot Experiments}

We evaluate real-world dual-arm closed-loop execution from two complementary perspectives: \emph{multimodal behavior preservation} and \emph{online inference efficiency}. All experiments share the robot and computing platform, using an NVIDIA RTX 4090 GPU for policy inference. Three tasks are considered: placement into either of two plates, bottle uprighting, and pen placement into a holder. The two-plate task directly tests preservation of spatially separated valid behavioral modes in real-world closed-loop execution; the other tasks compare success rates and online inference efficiency with $\pi_{0.5}$~\cite{physicalintelligence2025pi05}.

\subsubsection{Placement into Either of Two Plates: Multimodal Behavior Preservation}

In Fig.~\ref{fig:real-two}(a), the robot grasps an object and places it in either plate; both outcomes succeed. Instructions leave the plate unspecified, so identical observations and instructions admit two spatially separated executable modes. This directly tests CTP's multimodal behavior preservation in real-world closed-loop execution.

We collected 300 demonstration episodes, with 150 for each placement mode. As shown in Fig.~\ref{fig:real-two}(b), the two sets of trajectories converge to the respective target regions, forming a balanced, spatially separated bimodal trajectory distribution.
Across 50 real-world trials, CTP selects the left plate 31 times and the right plate 19 times ($62\%$ and $38\%$, respectively). All 50 placements succeed. No trial shows averaged trajectories converging between the targets. CTP thus preserves both valid modes in real-world closed-loop deployment, without collapsing to one mode or an intermediate compromise trajectory.

\subsubsection{Bottle Uprighting and Pen Placement: Task Success and Inference Efficiency}

We also compare CTP and $\pi_{0.5}$ on bottle uprighting and pen placement into a holder (Fig.~\ref{fig:real-robot-comparison}). The former requires grasping fallen bottles at different positions and orientations and restoring them upright; the latter requires accurately placing tabletop pens into a holder after grasping. These assess object reorientation and precise placement, respectively.
On bottle uprighting, both $\pi_{0.5}$ and CTP achieve a success rate of $95\%$. On pen placement, their success rates are $85\%$ and $83\%$, respectively. All experiments use the same robot platform and NVIDIA RTX 4090 GPU. CTP reduces per-call policy inference latency from $218.24$\,ms to $75.80$\,ms, corresponding to a $2.88\times$ speedup. The results show that CTP substantially reduces online policy inference overhead while maintaining real-world success rates comparable to $\pi_{0.5}$.


%% file: sections/libero_results.tex
\subsection{LIBERO: Transfer to Vision--Language--Action Models}
\label{sec:libero-results}

To test CTP in robot manipulation conditioned on vision and language, we integrate it into $\pi_{0.5}$ and evaluate the same checkpoint on all four LIBERO suites. This experiment examines whether pretrained vision-language representations continue to support multitask closed-loop manipulation when parallel trajectory peak prediction replaces iterative action generation.

\begin{table}[!t]
\caption{Comparison of results on LIBERO.}
\label{tab:libero-comparison}
\centering
\scriptsize
\setlength{\tabcolsep}{2pt}
\renewcommand{\arraystretch}{1.05}
\begin{tabular*}{\columnwidth}{@{\extracolsep{\fill}}lccccc@{}}
\toprule
Method & Spatial & Object & Goal & Long & Average \\
\midrule
OpenVLA\cite{kim2024openvla} & $84.7$ & $88.4$ & $79.2$ & $53.7$ & $76.50$ \\
$\pi_0$\cite{black2024pi0} & $96.8$ & $98.8$ & $95.8$ & $85.2$ & $94.15$ \\
LingBot-VA\cite{li2026lingbotva} & $98.5$ & $99.6$ & $97.2$ & $\mathbf{98.5}$ & $\mathbf{98.45}$ \\
Motus\cite{bi2025motus} & $96.8$ & $99.8$ & $96.6$ & $97.6$ & $97.70$ \\
Fast-WAM\cite{yuan2026fastwam} & $98.2$ & $\mathbf{100.0}$ & $97.0$ & $95.2$ & $97.60$ \\
\midrule
$\pi_{0.5}$\cite{physicalintelligence2025pi05} & $97.0$ & $99.0$ & $\mathbf{98.0}$ & $93.0$ & $96.75$ \\
\ours & $98.0$ & $\mathbf{100.0}$ & $\mathbf{98.0}$ & $93.0$ & $97.25$ \\
\bottomrule
\end{tabular*}
\end{table}

As shown in Table~\ref{tab:libero-comparison}, CTP achieves success rates of $98.0\%$, $100.0\%$, $98.0\%$, and $93.0\%$ on Spatial, Object, Goal, and Long, respectively, averaging $97.25\%$. This is close to Motus ($97.70\%$) and Fast-WAM ($97.60\%$), below LingBot-VA ($98.45\%$), and above the locally reevaluated $\pi_{0.5}$ baseline ($96.75\%$). Because these success rates do not come from strictly matched protocols, cross-model differences only contextualize overall performance. The experiment shows that structured single-pass action output can support LIBERO multitask closed-loop manipulation while retaining a large vision-language backbone.

\begin{figure}[!t]
\centering
\includegraphics[width=\columnwidth,trim=0 10bp 0 0,clip]{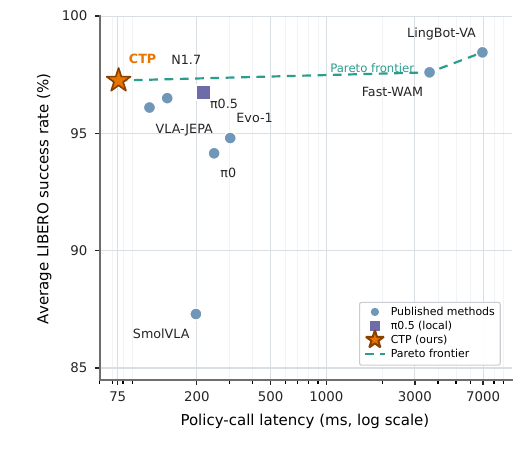}
\caption{Success--latency trade-off on LIBERO. CTP achieves an average success rate of $97.25\%$ with a mean policy-call latency of $75.80$ ms. All policy-call latencies were measured locally at batch size 1 on identical hardware comprising an NVIDIA GeForce RTX 4090 GPU and two Intel Xeon Gold 6530 CPUs. Success rates for published baselines are taken from their respective papers, whereas CTP and $\pi_{0.5}$ are evaluated under the same local protocol.}
\label{fig:libero-success-latency}
\end{figure}

CTP reduces mean latency for a complete policy invocation from $218.24$ ms to $75.80$ ms, a $2.88\times$ speedup and a $65.27\%$ latency reduction. P95 latency decreases from $219.64$ ms to $76.15$ ms. The local $\pi_{0.5}$ baseline uses $10$ flow integration steps per action chunk. CTP generates all trajectory peaks through one action expert forward pass, then selects a candidate using trajectory beliefs. Both predict $50$ steps and execute $10$, so the speedup does not result from a lower replanning frequency. Timing includes the vision-language backbone, data processing, and belief decoding. All policy-call latencies in Fig.~\ref{fig:libero-success-latency} were measured by us on the same local hardware. Success rates for published baselines are taken from their original papers and are therefore not protocol-matched; CTP and the local $\pi_{0.5}$ baseline use our unified evaluation.

%% file: sections/conclusion.tex
\section{Conclusion}

We presented CTP, a single-pass policy framework for multimodal closed-loop control. CTP models trajectory-level multimodal distributions in the space of complete action chunks. It combines DAPS and ETBT to promote peak specialization and maintain cross-chunk behavioral consistency during receding-horizon replanning, respectively. The framework thus integrates multimodal representation, closed-loop continuity, and single-pass inference. Experiments on Push-T, D3IL, and LIBERO show competitive multimodal closed-loop performance and substantially reduced online inference overhead compared with iterative action generation. Real-world experiments further demonstrate that CTP preserves multiple valid behavioral modes during closed-loop execution and achieves a $2.88\times$ policy inference speedup while maintaining task success rates comparable to $\pi_{0.5}$. Ablations confirm the complementary contributions of trajectory-level multimodal representation, DAPS, and ETBT. Overall, CTP provides a trajectory-level modeling approach to multimodal imitation learning that combines behavioral diversity, closed-loop consistency, and online efficiency.